\documentclass[sigconf ]{acmart}
\usepackage{graphicx}
\usepackage{dsfont}
\AtBeginDocument{%
  }

\setcopyright{acmcopyright}
\copyrightyear{2023}
\acmYear{2023}
\acmDOI{XXXXXXX.XXXXXXX}

\acmConference[CSAI]{International Conference on Computer Science and Artificial Intelligence}{December 8-10,
  2023}{Beijing, China}
\acmPrice{15.00}
\acmISBN{978-1-4503-XXXX-X/18/06}

\acmSubmissionID{AI-157-A}

\begin{document}

\title{Improve the efficiency of deep reinforcement learning through semantic exploration guided by natural language}

\author{Zhourui Guo}
\email{guozhourui2021@ia.ac.cn}
\orcid{}
\affiliation{%
  \institution{School of Artificial Intelligence, University of Chinese Academy of Sciences}
  \institution{Institute of Automation, Chinese Academy of Sciences}
  \city{Beijing}
  \country{China}
}

\author{Meng Yao}
\affiliation{%
  \institution{Institute of Automation, Chinese Academy of Sciences}
  \city{Beijing}
  \country{China}
  }
\email{meng.yao@ia.ac.cn}

\author{Yang Yu}
\affiliation{%
  \institution{School of Artificial Intelligence, University of Chinese Academy of Sciences}
  \institution{Institute of Automation, Chinese Academy of Sciences}
  \city{Beijing}
  \country{China}
}
\email{yuyang2019@ia.ac.cn}

\author{Qiyue Yin}
\affiliation{%
 \institution{School of Artificial Intelligence, University of Chinese Academy of Sciences}
  \institution{Institute of Automation, Chinese Academy of Sciences}
  \city{Beijing}
  \country{China}}
\email{qyyin@nlpr.ia.ac.cn}

\renewcommand{\shortauthors}{Guo et al.}

\begin{abstract}
Reinforcement learning is a powerful technique for learning from trial and error, but it often requires a large number of interactions to achieve good performance. In some domains, such as sparse-reward tasks, an oracle that can provide useful feedback or guidance to the agent during the learning process is really of great importance. 
However, querying the oracle too frequently may be costly or impractical, and the oracle may not always have a clear answer for every situation. Therefore, we propose a novel method for interacting with the oracle in a selective and efficient way, using a retrieval-based approach. 
We assume that the interaction can be modeled as a sequence of templated questions and answers, and that there is a large corpus of previous interactions available. We use a neural network to encode the current state of the agent and the oracle, and retrieve the most relevant question from the corpus to ask the oracle. 
We then use the oracle's answer to update the agent's policy and value function. We evaluate our method on an object manipulation task. We show that our method can significantly improve the efficiency of RL by reducing the number of interactions needed to reach a certain level of performance, compared to baselines that do not use the oracle or use it in a naive way.

\end{abstract}


\maketitle

\section{Introduction}
\begin{figure}[htbp]
  \centering
  \includegraphics[width=\linewidth]{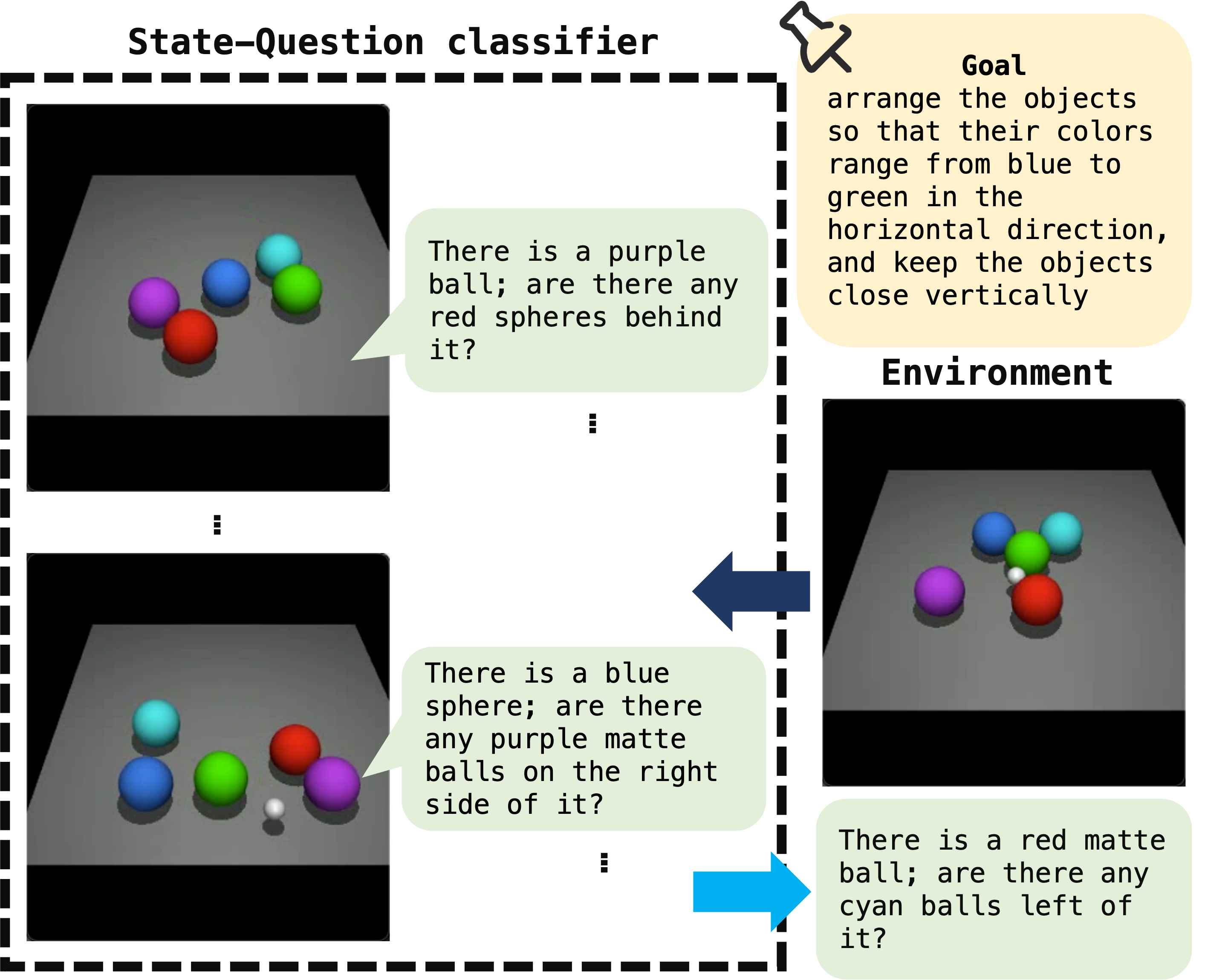}
  \caption{An overview of our approach. We proposes an efficient way to ask questions about the environment.}
\end{figure}
Reinforcement learning (RL) algorithms have demonstrated significant success in solving complex decision-making problems \cite{kaelbling1996reinforcement}. However,
RL agents often face complex and large-scale problems that require a lot of exploration and learning to find optimal policies \cite{luketina2019survey}. A possible way to enhance the performance of RL agents is to use human feedback or guidance during the learning process  \cite{goyal2021pixl2r,waytowich2019narration}. Incorporating human feedback or guidance is indeed a promising approach to address the challenges associated with the vast problem space in RL \cite{zhong2021silg}. Humans are able to seek help from external sources of knowledge, such as books, experts, or online resources, when they encounter difficult tasks. Humans also create sources of knowledge, such as manuals, tutorials, or demonstrations, because they can benefit from them in the long run. In this paper, we propose a novel approach to improve the efficiency of RL by incorporating an oracle during the interaction process. An oracle is a source of knowledge that can provide answers to the agent's queries. The agent can query the oracle when it is uncertain or stuck, and use the oracle's response to update its policy or behavior. We show that our approach can significantly improve the efficiency and performance of RL agents.

However, it also poses several challenges, such as how to elicit useful and timely feedback from humans \cite{nair2022learning}, how to integrate oracle feedback with other sources of information \cite{hermann2017grounded}, and how to ensure that the feedback is consistent and aligned with the agent's objectives \cite{lin2022inferring,waytowich2019grounding}. Moreover, the problem space generated in the interaction is large, and the interaction can often be templated \cite{kaur2021ask}, which means that there may be many similar or redundant questions that the agent can ask the oracle. To address this issue, we introduce a retrieval method that leverages templated interactions. By using a retrieval-based approach, we aim to identify and retrieve the most suitable questions or queries that can facilitate efficient learning.

In this paper, we present a new method that integrates RL with an oracle, which offers advice and feedback to the learner. We employ a retrieval technique to communicate and select the best questions. Using a retrieval method to select suitable questions from a database of past interactions is a practical approach. It can help the RL agent make informed decisions about what to ask the oracle, reducing unnecessary queries and potentially speeding up the learning process. Our method is motivated by the idea that the learner can benefit from past interactions and apply them in novel scenarios. We suppose that there is an oracle that can give optimal responses for any question, and that the learner can access a repository of previous interactions that contains question-answer pairs from the oracle. The learner can then choose the most suitable one in the repository to query the oracle based on their relation to the current state and action, which allows them to reuse knowledge and adapt to different situations. This way, the learner can prevent asking irrelevant or redundant questions, and concentrate on the most useful ones. By communicating with the oracle, the RL learner can obtain valuable information and expert knowledge, thus enhancing the learning process.

The contribution of this research lies in the design of a curiosity-driven intrinsic reward mechanism that effectively guides reinforcement learning agents to explore their environment more intelligently. By emphasizing actions that lead to substantial changes in the environment, our approach encourages the agent to discover novel and informative states, ultimately enhancing its learning performance and adaptability. We evaluate the effectiveness of our approach on CLEVR-Robot Environment \cite{jiang2019language}, and compare it with baseline methods that use random question selection. We show that our approach can significantly improve the efficiency and performance of RL by interacting with oracle, and that it can also reduce the cognitive load on the human. The retrieval-based approach allows the RL agent to focus on the most relevant and informative interactions, leading to faster convergence and better policy learning. We also analyze the properties and limitations of our approach, and discuss possible extensions and future directions.

Overall, our study contributes to enhancing the efficiency of reinforcement learning by leveraging an oracle during the interaction process. The proposed retrieval method enables the RL agent to interact effectively with the oracle, making use of templated questions and retrieving the most suitable queries. This research opens up new possibilities for improving the efficiency and scalability of RL algorithms in real-world applications.

\section{Background}

\subsection{Language for exploration}
Reinforcement learning is a branch of machine learning that deals with learning from actions and rewards \cite{kaelbling1996reinforcement}. A common framework for modeling reinforcement learning problems is the partially observable Markov Decision Process (POMDP). Markov Decision Processes (MDPs) are mathematical frameworks that capture the essential features of a reinforcement learning problem: states, actions, rewards, and transitions. States are the possible configurations of the agent and the environment. Actions are the choices that the agent can make in each state. Rewards are the immediate feedback that the agent receives after taking an action. Transitions are the probabilities of moving from one state to another after taking an action. MDPs assume that the environment is fully observable, meaning that the agent knows the current state at all times. A POMDP is a generalization of a MDP that accounts for the uncertainty in the agent's observations of the environment. In a POMDP, the agent does not have full access to the state of the environment, but only receives some partial or noisy observation. The agent's goal is to learn a policy that maximizes the expected cumulative reward over time, given the available observations and actions.

One of the main difficulties in RL is to find a trade-off between exploration and exploitation, that is, to gather sufficient information about the environment while optimizing the expected reward \cite{bahdanau2018learning}. Exploration refers to the agent's behavior of trying new actions or visiting new states, which may lead to discovering better policies in the long run. Exploitation refers to the agent's behavior of following the current best policy, which may lead to higher immediate rewards. Exploration can help the agent discover better actions and improve its performance, but it can also incur costs or risks. Exploitation can help the agent achieve immediate rewards, but it can also prevent the agent from finding better actions in the long run. A good RL agent should balance these two aspects, as focusing too much on either one may result in suboptimal performance. A possible research direction in reinforcement learning (RL) is to leverage natural language as a source of intrinsic motivation or curiosity for RL agents. Language can provide a natural and efficient way to guide exploration in RL, as it can encode rich and diverse information about the world, such as goals \cite{narasimhan2018grounding}, preferences, constraints \cite{hanjie2021grounding}, and feedback \cite{sumers2021learning}.
In this section, we review some of the existing work on using language for exploration in RL.

One line of work explores using natural language as a form of intrinsic motivation or curiosity for RL agents \cite{fu2019language}. This means that RL agents can use natural language inputs, such as text or speech, to generate their own goals and rewards, and to explore their environment in a self-supervised manner. By doing so, RL agents can learn from diverse and rich natural language signals, and acquire generalizable and transferable skills that can be applied to different tasks and domains. For example, some works \cite{mu2022improving, raileanu2020ride, campero2020learning, akakzia2020grounding} use natural language descriptions of the environment states as intrinsic rewards for exploration. The agent learns to generate and follow natural language plans that describe novel states or transitions. References \cite{kaur2021ask, carta2022eager, goyal2019using} use natural language questions as intrinsic rewards for exploration. The agent \cite{liu2022asking, chan2019actrce} learns to ask and answer questions that are informative and relevant to the task. Reference \cite{tam2022semantic} define novelty using semantically meaningful state abstractions, which can be found in learned representations shaped by natural language. A reward shaping approach \cite{mirchandani2021ella} geared towards boosting sample efficiency in sparse reward environments by correlating high-level instructions with simpler low-level constituents. Imaginative agent \cite{colas2020language} models ability of how children use language as a tool to imagine descriptions of outcomes they never experienced before and target them as goals during play.  These methods leverage language as a source of reward shaping, which can help the agent explore more efficiently and learn faster.

Another line of work investigates using natural language as a form of communication or coordination. This involves studying how natural language can facilitate human-computer interaction, human-human collaboration, and multi-agent systems \cite{zhong2019rtfm}. For example, natural language can enable users to interact with complex systems using natural commands or queries \cite{griffith2013policy, lazaridou2020emergent}, or to collaborate with other users or agents using natural dialogue or negotiation \cite{lowe2020interaction}. Natural language can also enable agents to coordinate their actions or goals using natural protocols or contracts \cite{yu2018interactive}. Some works \cite{gupta2021dynamic, lazaridou2020multi, tasrin2021influencing}use natural language messages as a communication channel between agents in a cooperative environment. The agents learn to generate and interpret messages that facilitate coordination and collaboration. Reference \cite{lewis2017deal} gathers a large dataset of human-human negotiations on a multi-issue bargaining task, where agents who cannot observe each other’s reward functions must reach an agreement (or a deal) via natural language dialogue. The agents learn to generate and understand dialogue acts that influence the outcome of the negotiation. These methods utilize language as a means of social interaction or persuasion, which can enable the agents to explore more complex and strategic behaviors. This research area aims to develop models and methods that can understand, generate, and reason with natural language in various domains and scenarios.

\subsection{CLEVR-Robot Environment}
CLEVR-Robot Environment \cite{jiang2019language} is a novel framework for developing and evaluating embodied agents that can perform complex tasks in 3D environments. It consists of a rich and diverse set of scenes, objects, and questions that can be used to measure the agent's perception, reasoning, and navigation skills. The environment is based on the CLEVR dataset \cite{johnson2017clevr}, which is a synthetic benchmark for visual question answering. CLEVR-Robot Environment extends CLEVR by adding realistic physics, lighting, and textures, as well as a controllable robot arm that can manipulate the objects in the scene. The agent can interact with the environment through natural language commands and queries, and receive feedback from a human-like instructor. The goal of CLEVR-Robot Environment is to provide a challenging and scalable testbed for embodied AI research. 
The environment can contain up to 5 objects with customizable colors, shapes, sizes and materials. There are 4 invisible planes on the table in 4 basic directions, which prevent the objects from leaving the table. The environment supports both state-based and image-based observations. It also contains a scene graph or world state that keeps track of the objects and their locations.

\section{Method}
\begin{figure*}[htbp]
\centerline{\includegraphics[width=1\textwidth]{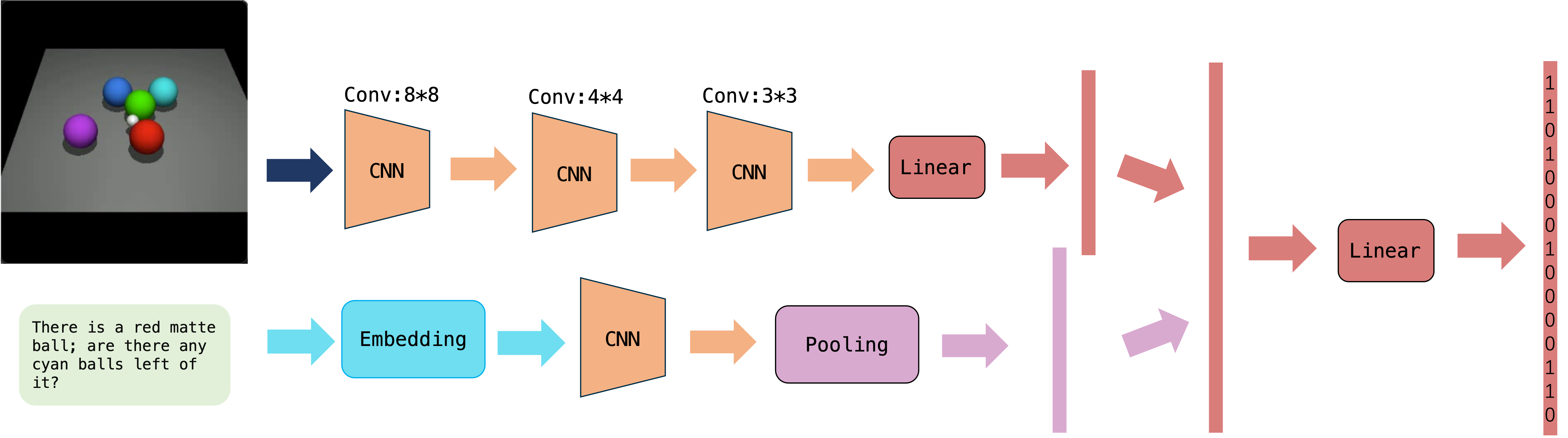}}
\caption{The architecture of our classifier.}
\label{fig}
\end{figure*}

One of the challenges of reinforcement learning is to design a suitable reward function that can guide the agent towards the desired behavior. Often, the reward is sparse. Alternatively, we can use intrinsic rewards that are generated by the agent itself based on its own internal state or curiosity. This way, the agent can use questions as a form of self-guidance and self-evaluation. To simplify the question generation process, we can use templates that cover most of the possible scenarios and types of questions that the agent might encounter. By using templates, we can avoid generating irrelevant or redundant questions and focus on the ones that are most informative and useful for the agent's learning. This can save time and effort, and ensure consistency and quality of the questions. By using intrinsic rewards and question generation, we can enable the agent to learn more effectively and efficiently from environment.

In this paper, we propose a novel approach to generate intrinsic rewards by using the answer of questions as a proxy for the agent's learning progress. We formulate the problem as a language guiding task, where the agent is given a set of questions related to its environment and its actions, and it can get an answer that is consistent with its observations. Our approach is based on pre-training a state-question map using supervised learning from a large data set that we collect beforehand. This map assigns a question to each state, such that the answer to the question provides a reward signal for the agent. We use this map to guide the exploration and learning of the agent in novel environments. For example, "There is a red matte ball; are there any cyan balls left of it?". The answer of each question is yes or no. Then used as an intrinsic reward signal that reflects how well the agent understands its surroundings and how effectively it can achieve its goals. We evaluate our method on CLEVR-Robot environment and show that it can improve the performance and generalization of the agent compared to baseline methods that use extrinsic rewards or other forms of intrinsic rewards.

To generate a curiosity-driven intrinsic reward for the agent, we first pretrained a classifier that can select the most relevant question for the current state, based on the magnitude of the intrinsic reward. The intrinsic reward is defined as the impact of an agent's action on the environment, which is calculated as the difference between the answers to the same question before and after the agent's action. This difference indicates how much the agent has altered the state of the environment by its action. The intrinsic reward can be used to motivate the agent to explore and learn from the environment, as well as to evaluate the agent's performance. We assume that questions with larger intrinsic rewards are more important, as they indicate that the agent has manipulated the objects in the environment. At each step, the agent queries the state-question map and retrieves the most similar question for the current state. The change in the answer to this question, \(A(s_{t}, q_k)\), after the agent's action, \(A(s_{t+1}, q_k)\), constitutes the curiosity signal for that action.

\subsection{Pretrained a classifier}
Our model uses both visual and linguistic inputs to produce a desired output, allowing the model to learn and generate meaningful representations that integrate information from both modalities.
The model consists of two main components: a visual encoder and a linguistic encoder. Visual Encoder is responsible for processing a sequence of frames, from image data, and extracting meaningful features from them. It uses three separate Convolutional Neural Network (CNN) feature extractors, each designed to capture different aspects of visual information. These extracted features are then concatenated into a single vector representation. Finally, this concatenated feature vector is passed through a linear layer, which transforms it into an encoding of the trajectory, summarizing the visual information. 
The linguistic encoder takes a given linguistic description as input. It begins by embedding the words or tokens in the description into a continuous vector space using an embedding layer. Then, the embedded representation is processed by a CNN, which is designed to capture the contextual and structural information in the text. The resulting features are flattened and passed through a linear layer to obtain an encoding of the linguistic description.
The decoder takes as input the concatenation of the visual and linguistic encodings, effectively combining information from both modalities. This combined encoding is passed through a linear layer, which generates the final output. The nature of the output will depend on the specific task or application of the model. For example, in some cases, it may generate a text description, while in others, it might produce action commands, predictions, or other relevant information.

The output vector consists of binary values (0 or 1) with a dimension of 256. Each element in the output vector corresponds to a specific question, and a value of 1 indicates that the question can potentially promote the reward, while a value of 0 indicates that the question does not have that potential. It allows the model to make decisions about which questions to ask or which actions to take based on their potential impact on maximizing the reward. By using binary values, our model can effectively select a subset of questions or actions that are most likely to lead to a favorable outcome while ignoring those that are less promising. This binary output vector is used to guide the agent's exploration, focusing on questions that have the potential to lead to greater intrinsic rewards, and thus encouraging more informed and impactful actions in the environment.

\subsection{Curiosity-driven Intrinsic Reward}
We propose a novel intrinsic motivation scheme that encourages the agent to explore the environment by asking and answering questions that are relevant to the dynamics of the world. The questions being asked are directly related to the environment, which means that the exploration is guided by the agent's perception of what information is most valuable for its learning and adaptation. This can lead to more efficient exploration. Providing an intrinsic reward when the agent observes a change in the answer to a previously asked question serves as a form of positive reinforcement for information-seeking behavior. It encourages the agent to continue exploring and asking questions. The reward signal from changes in answers can help the agent identify unexpected or novel events in the environment. This can lead to the discovery of new strategies or insights that might have otherwise been missed. This way, our approach has the potential to make exploration more efficient by focusing the agent's attention on areas of the environment that are most likely to yield valuable information for its learning objectives.

In our approach, we leverage a state-questions classifier to filter out the questions that are likely to increase the intrinsic reward of the agent.  Using a classifier to estimate the expected reward of questions for each state allows the agent to focus its exploration efforts on questions that are more likely to yield valuable information. This can lead to more efficient learning. The classifier is trained on a large corpus of question-answer pairs and state representations, and it predicts the expected reward for each question given a state. We then use the classifier to choose the questions according to their reward potential for each state. 
At time \(t\), we select \(n\) questions, the difference in the answer before the transition \( A\left ( s_{t },q_{i}   \right )\), and after the transition  \( A\left ( s_{t+1 },q_{i}   \right )\) contributes to the curiosity signal corresponding to that action, the intrinsic reward expressed as:

\[
r_{t}=\sum_{i=1}^{n} \mathds{1} \left [ A\left ( s_{t },q_{i}   \right )\ne  A\left ( s_{t+1 },q_{i}   \right ) \right ] \]

This way, we can focus on the questions that are most relevant and informative for the agent's learning process.

\section{Experiments}

In this section, we present the experimental setup, evaluation protocol, and our results on CLEVR-Robot Environment.

We designed the experiment to understand the following key question: Do actions that have the ability to change the environmental state more effectively on training agent? To answer different aspects of this question, we first evaluate our method in task. Then, we assess the impact of different types of basic language understanding on the intrinsic reward which is essential to understand the role of language. Finally, we investigate the effect of language feedback frequency on the output performance. Investigating the effect of language feedback frequency on output performance is valuable for optimizing our method. It allows us to fine-tune how often agents should ask questions, receive feedback, or modify their behavior based on linguistic cues. It can help strike a balance between language-driven exploration and other learning factors.

Unlike previous work \cite{kaur2021ask} that use 256 questions for each state, which is computationally expensive and prone to overfitting, we propose a novel method to pretrain a questions-states classifier. This way, we can reduce the number of questions required for each state and make the reinforcement learning more efficient and robust. By reducing the number of questions needed for each state, our approach can significantly decrease the computational burden of reinforcement learning. This makes it more feasible to apply our method to real-world tasks and large-scale environments. 
Our evaluation of the method on CLEVR-Robot Environment benchmarks and the demonstrated outperformance of state-of-the-art methods in terms of accuracy and speed indicate the practical effectiveness of our approach.

We present a classifier that can accurately predict the state of a question based on a large-scale experiment. Our experiment consists of 10,000 samples, where each sample is a pair of a state and its corresponding question. We train our classifier for 200 epochs, where each epoch consists of two phases: training and testing. In the training phase, we update the parameters of our classifier using 2570 batches of data, and we compute the loss function to measure the performance. In the testing phase, we evaluate the accuracy of our classifier using 318 batches of data, and we select the model with the highest accuracy as the final model. Our classifier achieves an accuracy of 97.47\% on the test data, and a loss of 0.130 on the train data, which demonstrates its effectiveness and robustness.

\textbf{Experimental Setup:} We aim to tackle the sparse reward problem in object alignment tasks, where the agent has to arrange balls by their color according to a given instruction. For instance, the instruction could be: "There is a blue rubber sphere; are there any green rubber spheres to the left of it?". The agent receives a +10 reward if it successfully orders the objects in this arrangement, and 0 otherwise. We choose a +10 reward instead of +1 to mitigate the extreme reward sparsity which hindered the agent's learning progress. We use PPO as the reinforcement learning algorithm for all our experiments and keep the hyperparameters constant. We run each algorithm three times with different random seeds without any optimization, and report the mean and standard deviation of the results over the three runs. For all experiments, we set the rollouts to 128 steps. We apply 3 epochs of optimization for each rollout with every method. The episode ends when the agent reaches the goal or exceeds the maximum number of steps. The agent receives a binary reward at the end of the episode, which is 1 if it places the objects according to the goal's spatial relation and 0 otherwise.

\textbf{Evaluation Protocol:}
To assess the performance of the policy, we conduct periodic evaluations throughout the training process. Every 50 updates of the model parameters, we run an evaluation phase that consists of 10000 episodes where the policy interacts with the environment. The evaluation phase does not affect the training phase, as we do not use the evaluation data for updating the model. We report the success rate of the policy.

\begin{figure}[htbp]
\centerline{\includegraphics[width=0.4\textwidth]{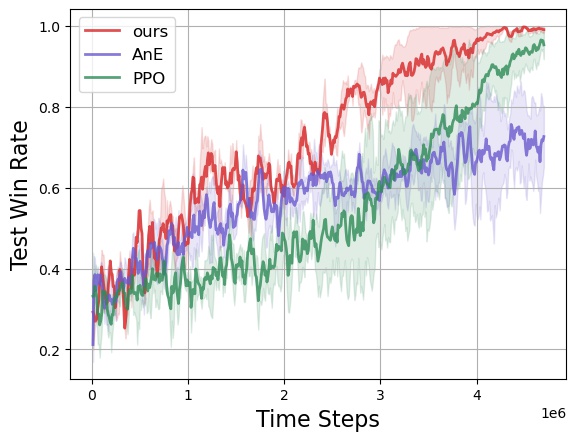}}
\caption{Our method significantly outperforms the baselines (AnE,PPO) which demonstrates the effectiveness of our language select method.}
\label{fig}
\end{figure}

\textbf{Results:}
As Fig3 demonstrates, the alignment task results indicate that our agent surpasses the performance of the baselines. This implies that our agent is capable of generating more pertinent and informative queries and utilizing the oracle feedback efficiently to accomplish the tasks.
The improved performance of our agent implies that it has a better ability to discern which questions are most relevant to the task at hand. This selective approach to questioning likely leads to more efficient and effective exploration.

\begin{figure}[htbp]
\centerline{\includegraphics[width=0.4\textwidth]{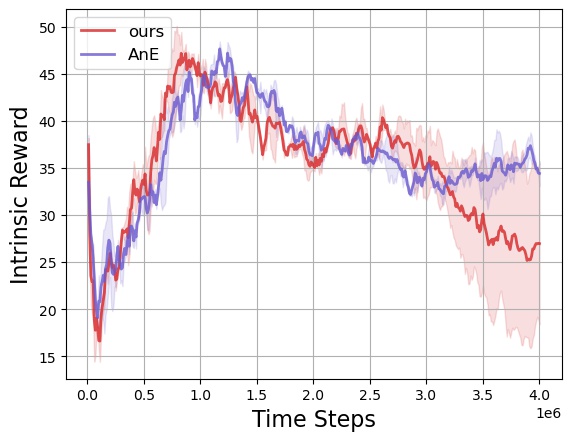}}
\caption{Our intrinsic reward compared with AnE.}
\label{fig}
\end{figure}

In addition, we observe the intrinsic reward in both our method and AnE method in Fig4. A large intrinsic reward at the beginning of RL model training indicates that the agents are highly motivated to explore the environment. This is often beneficial in the early stages of learning when the agent has limited knowledge about the environment. It helps the agent discover important information quickly. As the training progresses, both our method and the AnE method show a gradual decrease in intrinsic rewards. This suggests that the agents become more efficient at navigating the environment and rely less on curiosity-driven exploration. This behavior aligns with the idea that agents should transition from exploration to exploitation as they accumulate knowledge. 
Our model has faster convergence efficiency than the baseline implies that our approach accelerates the learning process, enabling the agent to learn optimal policies or strategies more quickly. 
The reduction in intrinsic rewards over time reflects the agent's adaptability and ability to shift its focus from exploration to exploitation as it gains experience. This adaptability is desirable as it allows the agent to strike a balance between exploration and exploitation throughout the learning process.

\textbf{Ablation Study:}
We conducted an ablation study to evaluate the impact of different steps of our method on the final performance. Specifically, we compared the following variants:

- Baseline: The method without any changes.

- Different questions: The method with inquiry 2 questions at each time and use average reward as the performance measure in order to investigate how the quantity of questions affects the success  of method.

- Different steps: The method with different steps to inquiry the oracle. The steps varied from 3 to 10, to investigate how the frequency of inquiry affects the success  of method.

We report the results in Fig5.

\begin{figure}[htbp]
\centerline{\includegraphics[width=0.4\textwidth]{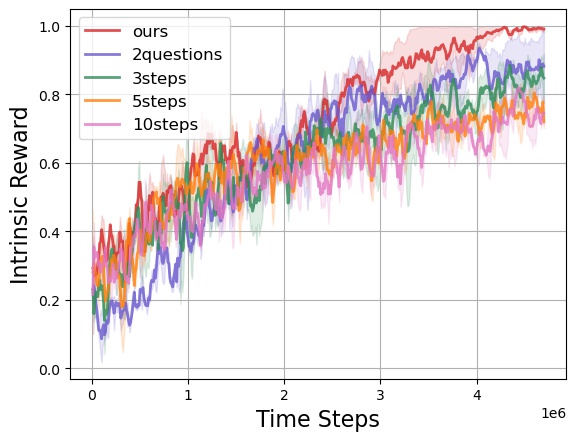}}
\caption{Ablation study of our method with different steps to inquire and the number of questions used. }
\label{fig}
\end{figure}

As shown in Fig5, 
the method that asks two questions has lower performance than asks one question, because it is hard to accurately describe the state with both questions. This complexity can lead to a decrease in the average intrinsic reward, which can not guide the agent well. The observation that asking questions with larger intervals leads to worse efficiency highlights the necessity of our method,  which involves optimizing the choice of questions and their frequency, strike a better balance between exploration and exploitation.

\section{CONCLUSION}
In this paper, we proposes a semantic exploration method based on natural language guidance, aiming to improve the efficiency of deep reinforcement learning for agents in complex environments. The method simulates the interaction between agents and humans, and integrates a templated way of posing questions, thereby overcoming the problem of large space of language interaction through retrieval. Specifically, this work designs a semantic retrieval model that dynamically retrieves questions according to the agent's state and action, and gives the agent corresponding rewards. In this way, the agent can adjust its strategy according to the feedback of the retrieved questions, and master the skills required for the task faster. The results show that this work has a significant advantage in improving the learning efficiency of agents, and can also achieve higher task performance. The reinforcement learning method based on retrieved questions proposed in this paper introduces more targeted guidance for the learning process of agents in complex environments, enabling them to learn and adapt more efficiently. This has important implications for enhancing the performance of agents in real-world applications, and also provides new insights for research in the field of reinforcement learning.

\bibliographystyle{ACM-Reference-Format}
\bibliography{bibliography}


\end{document}